\documentclass[preprints,article,accept,moreauthors,pdftex]{Definitions/mdpi}

\pdfoutput=1

\usepackage{tikz}
\usetikzlibrary{shapes,arrows}

\tikzstyle{arrow} = [thick,->,>=stealth]
\tikzstyle{line} = [draw, -latex']
\tikzstyle{cloud} = [draw, ellipse,fill=red!20, node distance=2.5cm, minimum height=1em]
\tikzstyle{decision} = [diamond, minimum width=3cm, minimum height=1cm, node distance=6cm, text centered, draw=black, fill=yellow!30]
\tikzstyle{process} = [rectangle, minimum width=3cm, text width=6cm, minimum height=1cm, node distance=3cm, text centered, draw=black, fill=orange!30]
\tikzstyle{io} = [trapezium, trapezium left angle=80, trapezium right angle=100, text width=6cm, node distance=6.5cm, minimum height=1cm, text centered, draw=black, fill=blue!30]   

\firstpage{1} 
\pubvolume{22}
\issuenum{6}
\articlenumber{2346}
\pubyear{2021}
\copyrightyear{2020}
\datereceived{24 February 2022} 
\dateaccepted{14 March 2022} 
\datepublished{18 March 2022} 
\hreflink{https://doi.org/10.3390/s22062346} 

\usepackage{pgfplots}
\usepackage[font=footnotesize]{caption}
\usepackage[font=footnotesize]{subcaption}
\usepackage[labelformat=simple]{subcaption}

\DeclareCaptionLabelFormat{subcaptionlabel}{\normalfont(\textbf{#2}\normalfont)}
\usepackage{algorithm}
\usepackage{array}
\usepackage{multirow}
\newcolumntype{P}[1]{>{\centering\arraybackslash}p{#1}}

\usepackage{verbatim}

\Title{Automated Feature Extraction on AsMap for Emotion Classification Using EEG}

\TitleCitation{Automated Feature Extraction on AsMap for Emotion Classification Using EEG}

\Author{Md. Zaved Iqubal Ahmed$^{1,*\href{https://orcid.org/0000-0002-8416-5819}{\includegraphics[width=.25cm]{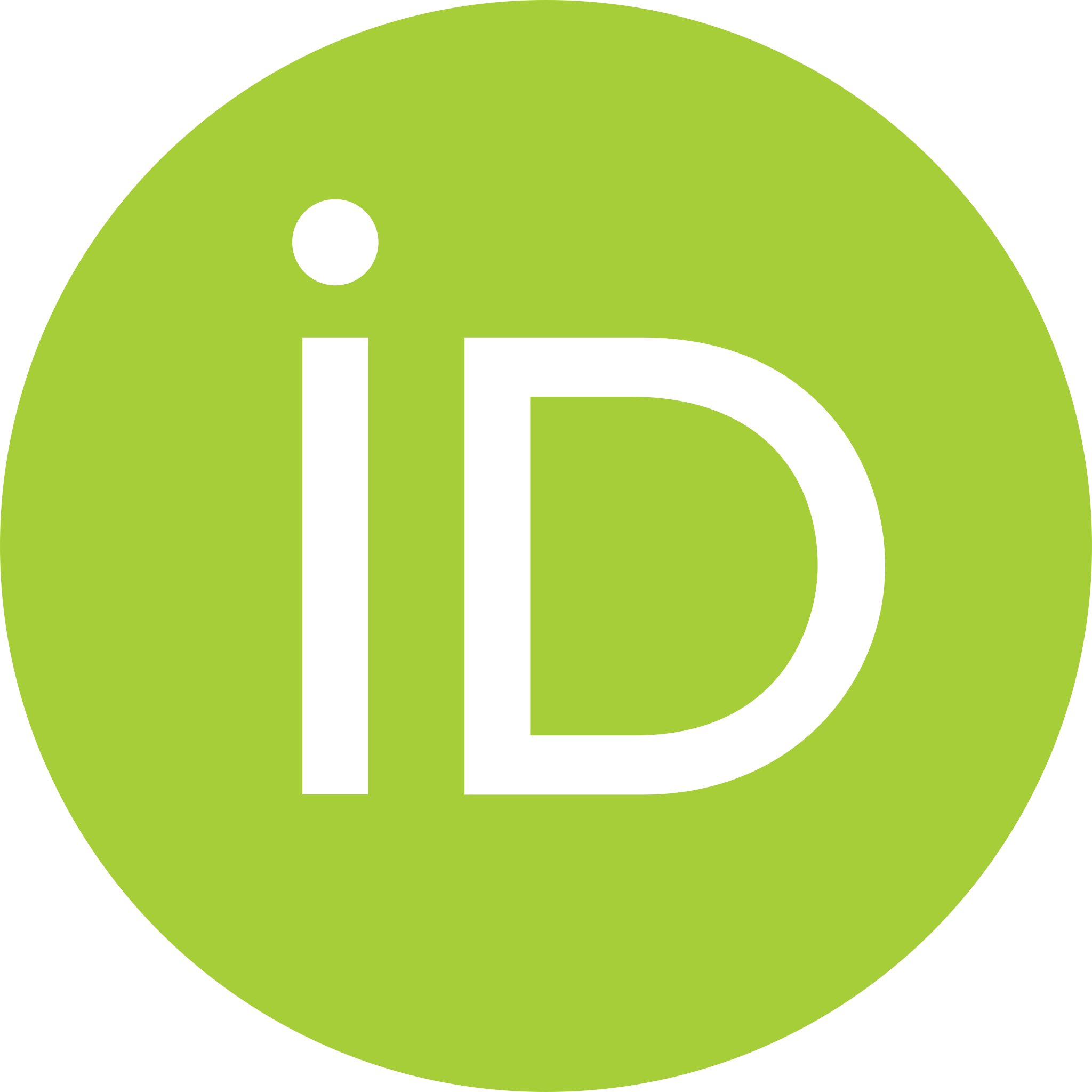}}}$, Nidul Sinha $^{2\href{https://orcid.org/0000-0003-0410-2154}{\includegraphics[width=.25cm]{Definitions/ORCID.png}}}$, Souvik Phadikar $^{2\href{https://orcid.org/0000-0002-7122-2095}{\includegraphics[width=.25cm]{Definitions/ORCID.png}}}$ and Ebrahim Ghaderpour $^{3,*\href{https://orcid.org/0000-0002-5165-1773}{\includegraphics[width=.25cm]{Definitions/ORCID.png}}}$}

\AuthorNames{Md. Zaved Iqubal Ahmed, Nidul Sinha, Souvik Phadikar and Ebrahim Ghaderpour}

\AuthorCitation{Ahmed, M.Z.I.; Sinha, N.; Phadikar, S.; Ghaderpour, E.}

\address{$^{1}$ \quad Department of Computer Science \& Engineering, National Institute of Technology, Silchar 788010, India\\
$^{2}$ \quad Department of Electrical Engineering, National Institute of Technology, Silchar 788010, India; \url{nidulsinha@ee.nits.ac.in} (N.S.); \url{souvik\_rs@ee.nits.ac.in} (S.P.)\\
$^{3}$ \quad Department of Geomatics Engineering, University of Calgary, Calgary, AB T2N 1N4, Canada}

\corres{Correspondence: \url{md.\_rs@cse.nits.ac.in} (M.Z.I.A.); \url{ebrahim.ghaderpour@ucalgary.ca} (E.G.)}

\abstract{Emotion recognition using EEG has been widely studied to address the challenges associated with affective computing. Using manual feature extraction methods on EEG signals results in sub-optimal performance by the learning models. With the advancements in deep learning as a tool for automated feature engineering, in this work, a hybrid of manual and automatic feature extraction methods has been proposed. The asymmetry in different brain regions is captured in a 2D vector, termed the AsMap, from the differential entropy features of EEG signals. These AsMaps are then used to extract features automatically using a convolutional neural network model. The proposed feature extraction method has been compared with differential entropy and other feature extraction methods such as relative asymmetry, differential asymmetry and differential caudality. Experiments are conducted using the SJTU emotion EEG dataset and the DEAP dataset on different classification problems based on the number of classes. Results obtained indicate that the proposed method of feature extraction results in higher classification accuracy, outperforming the other feature extraction methods. The highest classification accuracy of 97.10\% is achieved on a three-class classification problem using the SJTU emotion EEG dataset. Further, this work has also assessed the impact of window size on classification accuracy.}


\begin{document}


\section{Introduction}

Human emotions play a central role in decision making, social interaction, diagnosis of mental conditions such as depression, etc.~\cite{pan2020eeg,despande2017depression}. Traditionally, humans identify emotions using facial expressions, audio signals, body pose, gesture, etc.~\cite{wioleta2013using}. In contrast, machines cannot understand the feelings of an individual.
In this context, affective computing aims to improve communication among individuals and machines by recognizing human emotions, thus making this interaction more accessible, usable, and effective~\cite{rached2013emotion}.

Emotional experience is associated with physiological changes
in the body. Therefore, the knowledge of the physiological
reaction of every emotion is essential to emotion analysis~\cite{sheykhivand2020recognizing}. Thus, research works have been conducted to recognize emotions using physiological signals.
The physiological signals~\cite{bota2019review,shu2018review} are internal signals,
such as electroencephalogram (EEG), electrocardiogram, heart rate, electromyogram
(EMG), and galvanic skin response (GSR). According to Cannon’s theory~\cite{cannon1927james}, the emotion changes are
associated with quick responses in physiological signals coordinated by the autonomic nervous systems. This makes the physiological signals not easily controlled and overcomes the shortcomings of
bodily expressions~\cite{shu2018review}.

The advancement of brain--computer interface (BCI) devices and their ease of operation has motivated research on emotion recognition using EEG signals. Some of the non-invasive EEG devices are Emotiv Epoc, Emotiv Insight, Neurosky MindWave, InteraXon Muse, and OpenBCI. These devices are low-cost and portable, thus making EEG signals highly accessible. These devices are accompanied by tools for various BCI applications as well. The EEG signals are captured from individuals (or subjects) using the BCI devices and analyzed using computers to identify the emotion class. At the heart of emotion recognition lies the task of emotion classification. Emotion classification is the process of distinguishing one emotion from another. Emotions are categorized based on two types of models: categorical models and dimensional models. The categorical model categorizes emotions into discrete classes, commonly anger, disgust, fear, joy, sadness, and surprise~\cite{ps2017emotion}. Based on facial expression, Ekman listed six basic emotions: happiness, anger, fear, sadness, surprise, and disgust~\cite{ekman1992there}. On the other hand, the dimensional emotion model suggests that emotions can be placed in one or more dimensions rather than in categories. One of the popular dimensional models is the Circumplex model, where emotions are placed into two dimensions: valence (a continuum that varies from negative to positive) and arousal (a continuum that varies from low to high)~\cite{russell1980circumplex}.   

Considering the emotion models, various research works have been conducted to trigger emotional events using images, music, audio-visual cues, etc., and subsequently record the EEG signals from individuals. Some of the popular publicly available EEG datasets prepared by applying audio-visual stimuli are DEAP~\cite{koelstra2011deap} and SEED~\cite{seed}. The EEG signals from the datasets are used by machine learning models in order to learn how to classify different emotions. Traditional machine learning approaches such as support vector machine~\cite{liu2018real, li2009emotion, duan2013differential, lin2009eeg, ackermann2016eeg}, linear discriminant analysis~\cite{ramirez2012detecting, mehmood2017optimal}, quadratic discriminant analysis~\cite{petrantonakis2010emotion}, k-nearest neighbors~\cite{duan2013differential, ozerdem2017emotion, petrantonakis2010emotion, pham2012emotion}, Naïve Bayes~\cite{mehmood2017optimal}, feed-forward neural network~\cite{khosrowabadi2009affective}, deep belief network~\cite{zheng2015investigating}, multi-layer perceptron neural network~\cite{ozerdem2017emotion}, etc., are commonly used in EEG-based emotion classification.

In this context, raw time-domain EEG signals are very complex to be handled by the machine learning models as the signals are non-stationary and contaminated by artifacts. Some of the significant physiological artifacts in EEG signals are eye movement, muscle activity, and eye blinks. Various research works have been conducted to remove artifacts from EEG signals~\cite{jiang2019removal}. Recently, automatic artifact removal techniques have gained much popularity~\cite{phadikar2020automaticeeg,phadikar2020automaticeye}. 
After removal of artifacts, the most important task is feature extraction. Feature extraction methods are applied to reduce the complexity as well as the dimensionality of input data to the learning models. Features are commonly extracted from the delta, theta, alpha, beta, and gamma frequency bands. Some of the feature extraction methods available in the literature are the asymmetry measure ~\cite{duan2013differential}, power spectral density (PSD)~\cite{liu2018real}, differential entropy (DE)~\cite{duan2013differential}, wavelet transform~\cite{ozerdem2017emotion,ghaderpour2021just,ghaderpour2021survey},  higher-order crossings~\cite{petrantonakis2010emotion}, common spatial patterns~\cite{li2009emotion}, asymmetry index~\cite{petrantonakis2011novel}, differential asymmetry (DASM), relative asymmetry (RASM), and differential caudality (DCAU)~\cite{zheng2015investigating}. Most feature extraction methods are manual and the selection of an appropriate method for emotion classification is still a challenging task~\cite{phadikar2019survey}.

In recent years, research works on automatic feature extraction using deep learning models have been explored in various problems such as speech recognition, vision system, pattern recognition, etc.~\cite{liu2017survey}. 
Convolutional neural networks (CNNs) have shown tremendous capability in extracting spatial features from input data such as images, etc. 
Various research works ~\cite{yang2018emotion,wang2018emotionet,donmez2019emotion,keelawat2019spatiotemporal,chen2019accurate,moon2018convolutional} claim that deep learning models have shown their ability in emotion classification using EEG over traditional approaches. The authors in~\cite{yang2018emotion} proposed a feature extraction method that combines CNN and RNN. The CNN is used to extract spatial features and RNN is employed to extract temporal features. Both the feature vectors obtained from CNN and RNN are concatenated and given as input to the learning model. Classification accuracy of 90.80\% and 91.03\% was achieved for valence and arousal classification, respectively, on the DEAP dataset. In~\cite{wang2018emotionet}, raw EEG data are given as input to a CNN architecture having 3D convolution kernels. The automated features extracted using 3D-CNN result in arousal and valence classification accuracy of 73.1\% and 72.1\%, respectively, on the DEAP dataset. Moon et al. in~\cite{moon2018convolutional} proposed a CNN-based approach for automated feature extraction. Three connectivity features, namely the Pearson correlation
coefficient, phase-locking value, and
phase lag index, are used to measure the cross-electrode relationship. Each connectivity feature is transformed into a 2D vector and given as input to different CNN models, such as CNN-2, CNN-5, and CNN-10, for automated feature extraction. The authors claimed accuracy of 99.72\% for valence classification on the DEAP dataset using CNN-5 with phase-locking value matrices. The authors in~\cite{chen2019accurate} proposed an automated emotion classification method using the CNN model on time-domain and frequency-domain features. 
 
In this work, a novel feature extraction method for emotion classification has been proposed. The EEG signals are first segmented into segments of fixed window size, and on each segment, DE features are calculated on five frequency bands. The method then generates a 2D feature map, termed the asymmetric map (AsMap), from the DE features obtained from an EEG segment. The AsMap features are then fed into a CNN for automated feature learning. The DE features give a measure of the randomness in the EEG signal. The DE of an EEG segment is considered to be equivalent to the logarithm energy spectrum of a specific frequency band~\cite{duan2013differential}. The mathematical aspects of DE have been further discussed in Section \ref{manual_ext}. Other feature extraction methods such as DASM, RASM, and DCAU are derived from DE features. DASM is the difference in DE features on channels between two brain hemispheres. On the other hand, RASM is the ratio in DE features on channels between two brain hemispheres. In DCAU, the difference between the DE features on frontal and posterior brain regions is calculated. However, the AsMap represents the difference between DE features between every channel pair in a 2D vector. Thus, capturing all the possible inter-channel asymmetry in the spatial domain results in more discriminating features compared to other methods such as DASM, RASM, etc. Further, the windowing/segmentation process also provides time-domain resolution for each AsMap. Thus, the AsMap captures both temporal as well as spatial features from all brain regions. The proposed method has been tested on the SEED as well as on the DEAP dataset and compared with other features such as DE, DASM, RASM, and DCAU. Different classification scenarios have been tested on the proposed method. 

The rest of the paper is organized as follows.   In Section \ref{sec:matherialsAndMethods}, the materials and methods used in automated feature extraction for emotion classification using the AsMap are discussed. Later, in Section \ref{sec:results}, the results obtained during the experiment are presented. Section \ref{sec:discussion} provides a discussion of the contributions and the limitations of the proposed method. 
Lastly, Section \ref{sec:conclusion} gives the conclusions and future work.


\section{Materials and Methods}
\label{sec:matherialsAndMethods}
\subsection{Public Datasets}
\subsubsection{SJTU Emotion EEG Dataset (SEED)}

Zheng et al.~\cite{zheng2015investigating} prepared an EEG emotion dataset in the Center for Brain-Like Computing and Machine Intelligence Laboratory by recording EEG signals. At the same time, participants were subjected to audio-visual stimuli. A total of 15 participants, comprising 7~males and 8 females, were part of the experiment. The SEED dataset considers three basic human emotions named positive, negative, and neutral. Positive emotion describes a pleasant or desirable state of mind, ranging from interest to contentment. On the other hand, a negative emotion depicts an unpleasant or unhappy state. Finally, the neutral emotion is associated with the feeling of indifference, nothing in particular, and a lack of preference. These emotions were elicited using 15 Chinese movie clips of length of around \mbox{4 min.} Each trial of the experiment had 5 s indicating the start, followed by the presentation of the movie clip. After completion of the movie, each participant was allotted 45 s for their self-assessment, and lastly, a 5 s resting time was provided. The self-assessment involved the following questions: (1) what did they feel after watching the movie clip? (2)~is he/she familiar with the movie clip? (3) have they understood the movie clip?

\textls[-25]{The EEG signals were captured using 62 electrodes placed according to the \mbox{10--20 system}.} The SEED dataset contains two parts: the first part contains the processed EEG recordings and the second part contains some extracted features. In the first part, the EEG recordings are down-sampled to 200 Hz, and EEG recordings containing artifacts such as EOG and EMG were visually checked. 
The recordings seriously contaminated by EMG and EOG were removed manually.  In order to filter the noise and remove the artifacts, a bandpass frequency filter from 0.3 to 50.0 Hz was applied. The dataset includes only the EEG captured while watching the movie clip, with the rest eliminated. For the second part, each channel of the EEG data was divided into same-length epochs of 1 s without overlapping. There were around 3300 clean epochs for one experiment. Features such as PSD, DE, DASM, RASM, and DCAU were computed on each epoch of the EEG data. The dimensions of PSD, DE, DASM, RASM, and DCAU features obtained were 310, 310, 135, 135, and 115, respectively. In order to further filter out irrelevant components, each feature vector was further smoothed using conventional moving averages and linear dynamic systems, which are then provided as separate feature vectors. 

One of the limitations of the SEED is that it was prepared on very few participants. Moreover,  the annotation of the video clips with emotion classes was not done by the participants. Thus, the participants' assessments after watching the videos were not considered for annotation in this dataset.     

\subsubsection{Database for Emotion Analysis Using Physiological Signals (DEAP)}
Sander Koelstra et al.~\cite{koelstra2011deap} prepared a multimodal dataset called DEAP containing EEG and physiological signals. The dataset was prepared from the recordings of 32 participants aged between 19 and 37 and had a balanced male--female ratio. Each participant was presented with 40 videos having emotional content. The 40 videos were selected out of \mbox{120 music} videos, which were collected from the website last.fm, having affective tags and a manual procedure. The selection procedure for the videos involved a web-based subjective emotion assessment interface. All the videos were of 1-min length and contained music videos. EEG was recorded at a sampling rate of 512 Hz using 32 active AgCl electrodes (placed according to the international 10--20 system). Thirteen peripheral physiological signals, such as GSR, respiration amplitude, skin temperature, electrocardiogram, blood volume by plethysmograph, electromyograms of Zygomaticus and Trapezius muscles, and electrooculogram (EOG), etc., were also recorded. 

The synchronization of the EEG with emotion data was done by first displaying a  fixation cross on the screen and asking the participant to relax for 2 min. After that, \mbox{40 videos} of 1-min length were presented in trials to each participant, and before each trial, a 2-s screen displayed the progress, and then a 5-s fixation cross was displayed to relax the participant. It is very difficult to find markers in EEG signals for transition status in emotions, as the transition status is highly subjective in nature. Therefore, the participant ratings were used to mark the induced emotion.   

The DEAP dataset contains the processed EEG recordings, which were further downsampled to 128 Hz, and the eye blink artifact was removed using blind source separation. A bandpass frequency filter from 4.0 to 45.0 Hz was also applied. The data were averaged to the common reference and they were segmented into 60-s trials and a 3-s pre-trial baseline (out of the 5-s baseline recording). Moreover, the participant ratings were supplied separately for valence, arousal, and dominance.

DEAP and SEED are the two most popular publicly available EEG emotion datasets. Both the datasets used audio-visual stimuli for emotion elicitation. The DEAP dataset has a greater number of EEG recordings compared to the SEED dataset as the numbers of participants and videos are higher than in the SEED dataset. Unlike the SEED dataset, the DEAP dataset recorded physiological signals apart from the EEG. However, the EEG recordings of the SEED dataset have higher spatial resolution compared to the DEAP dataset, as a higher number of electrodes were used in the SEED dataset to capture EEG signals. The DEAP dataset used 40 different 1-min video clips to induce emotion in the participants but SEED used 15 different movie clips of a maximum duration of 4 min. Lastly, the SEED dataset used a categorical emotion model, whereas the DEAP dataset used a dimensional emotion model. The proposed feature extraction method was experimented on both the datasets. 

\subsection{Proposed Methodology}

This section discusses the methodology behind applying the deep learning technique for automated feature learning from EEG data for emotion classification. The method involves three steps as given below:
\begin{itemize}
    \item Manual Feature Extraction;
    \item Generation of Asymmetric Map;
    \item Automated Feature Extraction.
\end{itemize}

\subsubsection{Manual Feature Extraction} 
\label{manual_ext}
As EEG signals are complex and non-stationary, introducing EEG signals directly for automated feature learning can lead to sub-optimal performance. 
Therefore, in this work, DE features are extracted from the EEG signals. Considering an EEG signal from a channel as a continuous random variable, DE gives the measure of the randomness in the EEG signal. The DE of an EEG segment is considered to be equivalent to the logarithm energy spectrum of a specific frequency band~\cite{shi2013differential}. The DE equation on a random variable is \mbox{given as }
\begin{linenomath}
\begin{equation}
    h(X)=-\int_{X}f(x)\log(f(x))dx
\end{equation}
\end{linenomath}

To extract the DE features, the frequency spectrum of an EEG signal in a channel is first obtained using a 256-point short-time Fourier transform (STFT) with a non-overlapping Hanning window of 1 s. 
 As different frequency ranges in EEG signals resemble different brain states, various research works pre-dominantly subdivide the waveforms into frequency bands such as delta, theta, alpha, beta, and gamma. Frequencies ranging from 1 Hz to 3 Hz are named the delta band, which indicates a sleep state. The theta band comprises frequencies ranging from 4 to 7 Hz and resembles a deeply relaxed state. The frequency band 8 to 13 Hz is named the alpha band and indicates a very relaxed and passive attention state. The beta band, comprising frequencies ranging from 14 to 30 Hz, resembles anxiety, external attention, and an active state. Frequencies ranging from 31 to 50 Hz, named the gamma band, represent a state of concentration and focus. The difference in the frequency ranges at low and high frequency is attributed to the rhythmic patterns associated with the brain states. 
 The DE features are extracted for each frequency band in every epoch, thus retaining the temporal characteristics. The DE features are further smoothed using moving average in order to eliminate any unintended component introduced in the features. Figure~\ref{fig:two_figure} (top panel) gives a pictorial representation of the manual feature extraction process.
 
 \subsubsection{Generation of Asymmetric Map}
After manual feature extraction, the next important step is to generate the AsMap. Previous works have shown that the asymmetrical brain activity seems to be effective in discriminating EEG signals induced by different emotions~\cite{davidson1982asymmetrical,wu2010optimal}. Here, the DE features of each frequency band in n consecutive epochs in an EEG segment are grouped in fixed-sized, non-overlapping windows, and we average the DE features under a window to form a vector of size m. As there are 62 channels, we obtain a $62 \times m$ vector for each frequency band. Each column in the 2D vector further undergoes transformation to generate an AsMap on the $k${th} frequency band using Equation (\ref{eqn2}). 
\begin{linenomath}
\begin{equation}
\label{eqn2}
    AsMap(i,j,k)= DE(i,k)-DE(j,k)   
\end{equation}
\end{linenomath}

\begin{figure}[H]%

     \begin{subfigure}[b]{0.45\textwidth}
         \captionsetup{justification=centering}
         \includegraphics[width=\textwidth]{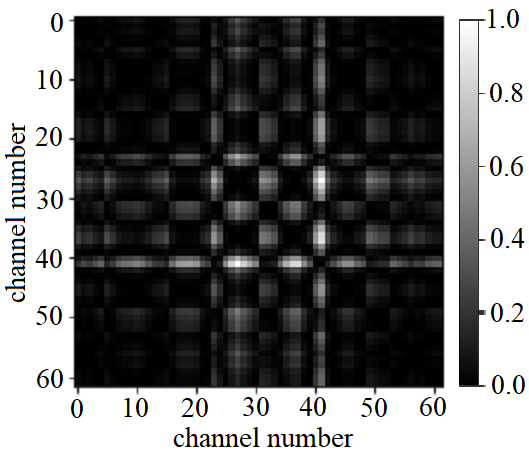}
         \label{fig:figure1}
     \end{subfigure}
     \hspace{1cm}
     \begin{subfigure}[b]{0.45\textwidth}
      \captionsetup{justification=centering}
         \includegraphics[width=\textwidth]{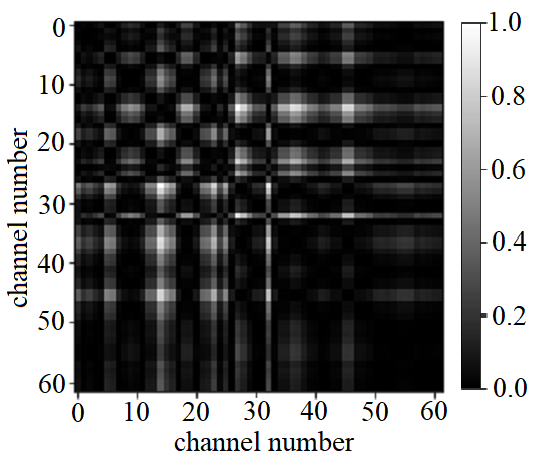}
         \label{fig:figure2}
     \end{subfigure}
     
     \begin{subfigure}[b]{0.45\textwidth}
 \captionsetup{justification=centering}
         \vspace{0.2cm}
         \includegraphics[width=\textwidth]{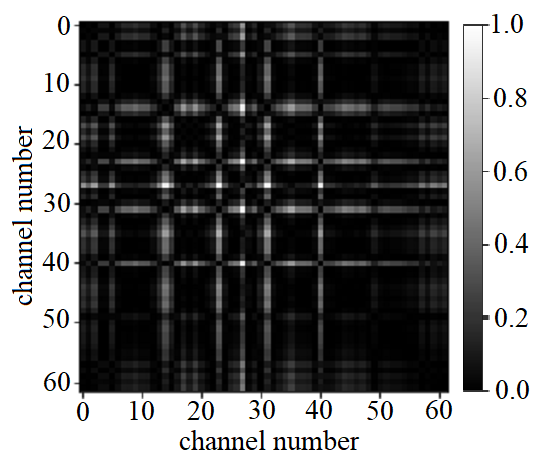}
\centering
         \label{fig:figure3}
     \end{subfigure}
        \caption{AsMap 
 of gamma band on a slot corresponding to positive, negative, and neutral emotion respectively.}
        \label{fig:AsMap}
\end{figure}

Here, $DE(i,k)$ represents DE features on the $k${th} frequency band of the $i${th} channel and $DE(j,k)$ represents DE features on the $k${th} frequency band of the $j${th} channel.
\vspace{-6pt}
\end{paracol}
\begin{figure}[H]

     \begin{subfigure}[b]{.97\linewidth}
        \captionsetup{justification=centering}
         \includegraphics[width=\linewidth]{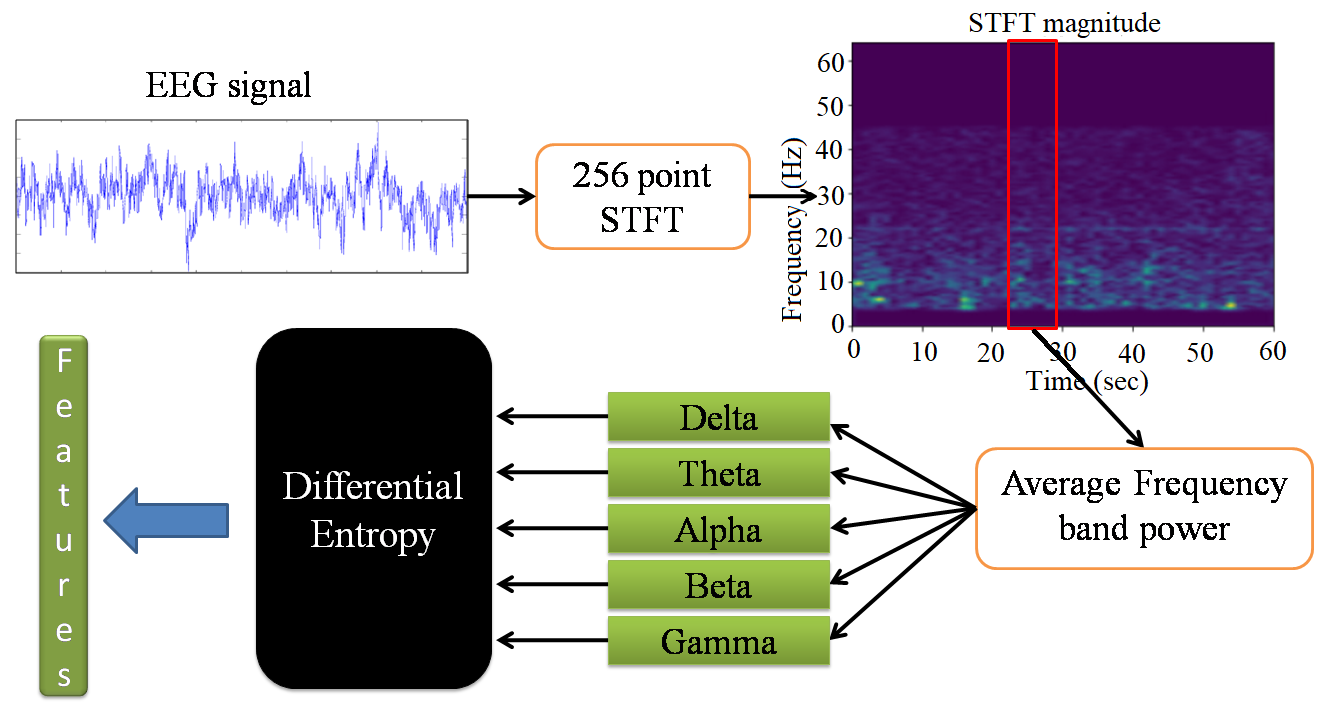}
         \label{fig:manual_extraction}
     \end{subfigure}
     \vspace{0.4cm}
%
%
%

     
     \begin{subfigure}[b]{.97\linewidth}
  \captionsetup{justification=centering}
         \includegraphics[width=0.96\linewidth]{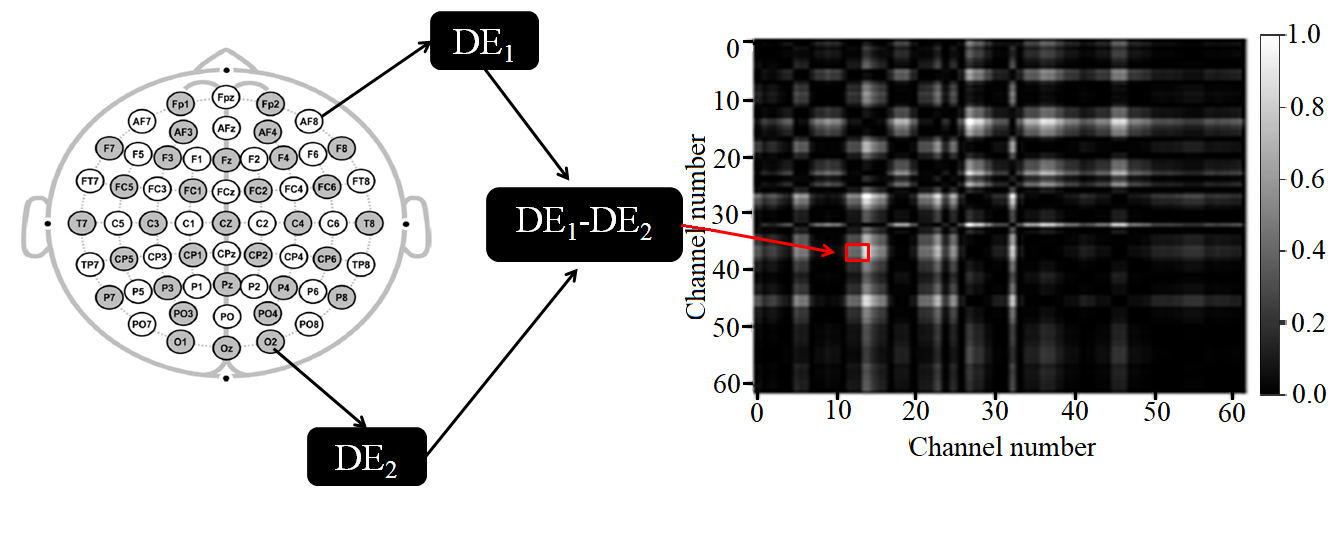}
         \label{fig:asmap_step}
     \end{subfigure}
     
        \caption{Pictorial 
 representation of the steps involved in manual feature extraction and generation of AsMap.}
        \label{fig:two_figure}
\end{figure}
\begin{paracol}{2}
\switchcolumn

Normalization is also performed on the AsMap to transform the data in such a way that each AsMap has distributions in a common scale from 0 to 1. The AsMap captures the difference in DE between all possible pairs of channels, as shown in Figure~\ref{fig:two_figure} (bottom panel). In the AsMap, the difference in DE features among all channel pairs gives a quantitative measure of the low-level asymmetry in different brain regions irrespective of their spatial location. For illustration, the AsMap of the gamma band for a slot in an EEG segment corresponding to positive, negative, and neutral emotion in the SEED dataset is presented as grayscale images in the top left, top right, and bottom left panels of Figure~\ref{fig:AsMap}, respectively.
\end{paracol}
\begin{paracol}{2}
\switchcolumn

\subsubsection{Automated Feature Extraction}
After obtaining the AsMap, we perform automated feature extraction on AsMaps of a subset of frequency bands to obtain patterns in the asymmetry of different brain regions across frequency bands. For this purpose, we use CNN on a subset of AsMaps to obtain a 1D feature vector. The CNN model has two 2D convolutional layers with a kernel size of \mbox{3 × 3} for spatial feature extraction. Further, each convolution layer uses the rectified linear unit (ReLU) activation function. The use of the 3 × 3 kernel and ReLU activation in this work is inspired by various models in the computer vision field. Initially, the first convolutional layer has 32 feature maps, but in the subsequent convolutional layer, the feature maps are halved to 16 feature maps. After each convolutional layer, we have a max pooling layer that strides a two-dimensional filter of size (2 × 2) over each channel of the feature maps and calculates the maximum or largest of the features lying within the region covered by the filter. It reduces the dimensions of the feature maps generated in the convolutional layer. The max pooling layer is followed by a dropout layer, where we randomly shut down 25\% of a layer’s neurons at each training step by zeroing out the neuron values. Finally, the feature maps from the last max pooling layer are flattened to obtain a 1D feature vector.   Different layers of the CNN model used in this work are presented in Figure~\ref{fig:model}.

\begin{figure}[H]
    \centering
    \includegraphics[width=200px,height=400px]{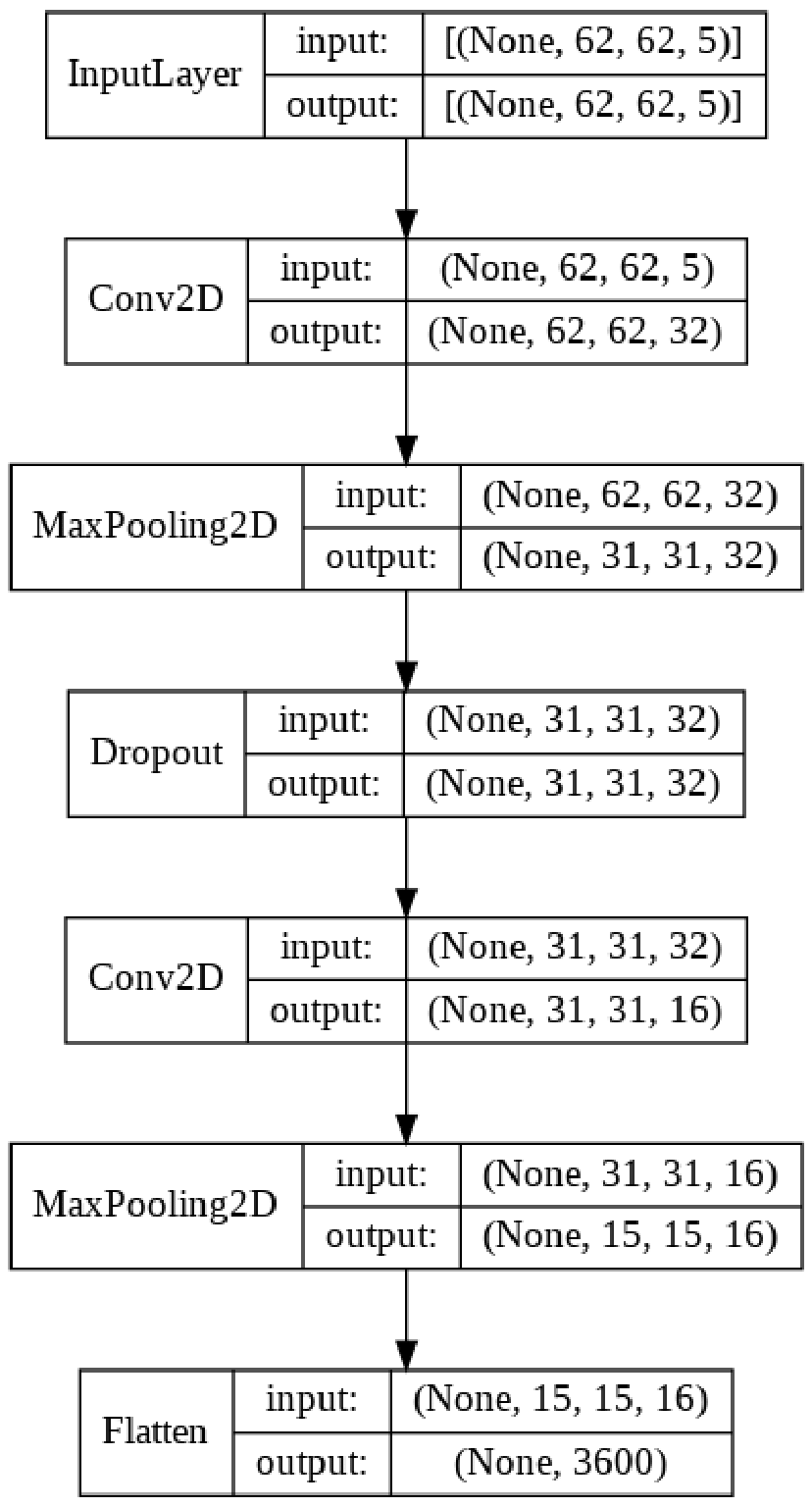}
    \caption{Different layers in the CNN model.}
    \label{fig:model}
\end{figure}

\section{Results}
\label{sec:results}
\subsection{Experimental Setup}
During the experiment, an Acer Desktop with Intel Core i3 $7${th} gen processor and \mbox{4GB RAM} was used. Anaconda 3, which is a free and open-source distribution of the Python and R programming language, was used to perform the scientific computing. Python libraries such as Numpy, Pandas, and Scikit-Learn are some of the most important libraries used for data handling during the experimentation. The proposed method for feature extraction was tested on both the SEED and DEAP datasets. The experiment conducted on the SEED dataset used the pre-extracted DE features. The DE features were used to generate the AsMap. As EEG recording in the SEED dataset contains signals from \mbox{62 channels,} the dimension of the AsMap is $62\times62\times k$ for all frequency bands together. Here, $k$ is the number of frequency bands. As the SEED dataset presents three classes of emotion (positive, negative, and neutral), a three-class classification problem on the SEED dataset was formulated. The classification problem was formulated to classify between positive, negative, and neutral emotions.
Further, experiments were conducted on the DEAP dataset, and AsMap features were extracted from the 32-channel EEG recordings. The dimension of the AsMap features extracted from the DEAP dataset was $32\times32\times k$  for all frequency bands together. Based on the valence and arousal ratings provided in the DEAP dataset, two different classification problems were formulated: two-class classification (valence classification and arousal classification) and four-class classification. The two-class classification on valence was to classify between high valence and low valence. Meanwhile, the two-class classification on arousal was to classify between high arousal and low arousal. During the preparation of the DEAP dataset, participants provided a rating from 1 to 9 for valence and arousal after watching each video. Based on the distribution of the subjective ratings~\cite{koelstra2011deap}, these ratings were considered as an estimate for valence and arousal. The classes were obtained in the following manner: the participants' ratings from 5.5 to 9 were categorized as the high-valence (HV) class and ratings from 1 to 5.5 were categorized as the low-valence (LV) class. Similarly, the participants' ratings from 5.5 to 9 were categorized as the high-arousal (HA) class and ratings from 1 to 5.5 were categorized as the low-arousal (LA) class.  In the four-class classification problem, both valence and arousal classes were combined together to classify four different classes of emotion.
The class labels for the four-class classification problem were high valence--high arousal (HVHA), high valence--low arousal (HVLA), low valence--high arousal (LVHA), and low valence--low arousal (LVLA).   

The 1D feature vector obtained in the automated feature learning process was used to train a fully connected neural network having two hidden layers with 512 neurons. Each hidden layer used the ReLU activation function. The output layer had a number of neurons equal to the number of classes, and the softmax activation function was used to classify the different classes of emotion. For comparison, other feature extraction methods such as DE, DASM, RASM, and DCAU were also used to train the classifier separately.

In order to analyze the proposed method on both the SEED and DEAP datasets, the classification accuracy using AsMap+CNN features was compared with DE and other DE-based features such as DASM, RASM, and DCAU. The features were obtained on different frequency bands such as delta ($\delta$), theta ($\theta$), alpha ($\alpha$), beta ($\beta$), gamma ($\gamma$), and all frequency bands together (ALL BAND).  Experiments were also conducted on varying window sizes, where the window size was set to 3 s, 6 s, 12 s, 30 s, respectively.

\subsection{Three-Class Classification on SEED}
Table~\ref{table1} presents the three-class emotion classification accuracy using different feature extraction methods such as DE, DASM, RASM, DCAU, and AsMap+CNN on delta ($\delta$), theta ($\theta$), alpha ($\alpha$), beta ($\beta$), gamma ($\gamma$), and all frequency bands together (ALL BAND).  The proposed method outperformed all the DE-based feature extraction methods on delta ($\delta$), theta ($\theta$), alpha ($\alpha$), beta ($\beta$), gamma ($\gamma$), and all frequency bands together (ALL BAND).  The highest classification accuracy of 97.10\% was obtained using AsMap+CNN on the $\gamma$ band with the use of a 3-s window size. It was also observed that the classification accuracy obtained using all the other feature extraction methods remained between 93\% and 96\% on the $\gamma$ band. Further, the features on $\beta$ and ALL BAND from all the feature extraction methods resulted in classification accuracy above 91\%, except for DE and RASM. It was also observed that the classification accuracy using different feature extraction methods on  delta ($\delta$), theta ($\theta$), and  alpha($\alpha$) remained below 70\%.
\end{paracol}
 \begin{table}[H] 
\caption{Three-class classification accuracy obtained using different feature extraction techniques on  frequency bands.
\label{table1}}
\newcolumntype{C}{>{\centering\arraybackslash}X}
\begin{tabularx}{\textwidth}{CCCCCCC}
\toprule
\textbf{Method} & \boldmath{$\delta$} & \boldmath{$\theta$} & \boldmath{$\alpha$} & \boldmath{$\beta$} & \boldmath{$\gamma$} & \textbf{ALLBAND} \\
\midrule
DE & 60.80\% & 47.41\% & 57.07\% & 88.09\% & 95.09\% & 88.28\% \\
RASM & 53.07\% & 49.56\% & 60.49\% & 88.53\% & 93.12\% & 90.62\% \\
DCAU & 59.79\% & 55.15\% & 64.02\% & 91.31\% & 95.12\% & 94.70\% \\
DASM & 57.44\% & 52.54\% & 63.58\% & 91.41\% & 95.87\% & 94.34\% \\
AsMap+CNN & 62.18\% & 56.20\% & 69.56\% & 93.99\% & 97.10\% & 96.25\% \\
\bottomrule
\end{tabularx}
\noindent{\footnotesize{ The window size was set to 3 s.}}
\end{table}
\begin{paracol}{2}
\switchcolumn
 The classification accuracy using AsMap+CNN on different frequency bands and window sizes is presented in Figure~\ref{fig:res1}. It can be observed that an increase in window size has a negative impact on the classification accuracy. Using AsMap+CNN features on $\beta$, $\gamma$, and ALL BAND, the classification accuracy remained above 85\% for window sizes smaller than or equal to 12 s. The classification accuracy obtained from features calculated on $\gamma$, $\beta$, and ALL BAND showed linear degradation, and the accuracy remained above 75\% until a 30 s window size. However, features obtained on delta ($\delta$), theta ($\theta$), and alpha ($\alpha$) did not show a linear degradation in accuracy. The figure also clearly illustrates that features on $\gamma$, $\beta$, and ALL BAND had greater discriminating ability than those of other bands. 
 \vspace{-6pt}
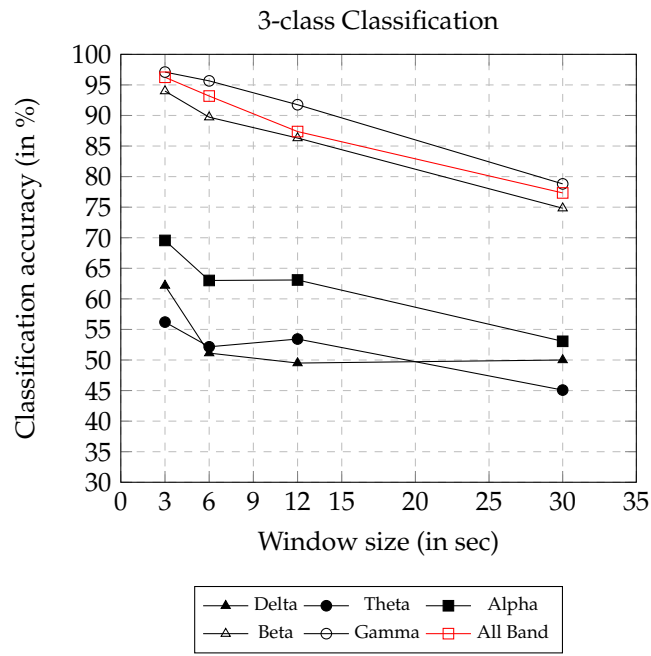
\begin{figure}[H]
\centering
    \begin{tikzpicture}
\begin{axis}[
    title={3-class Classification},
    xlabel={Window size (in sec)},
    ylabel={Classification accuracy (in \%)},
    xmin=0, xmax=35,
    ymin=30, ymax=100,
    xtick={0,3,6,9,12,15,20,25,30,35},
    ytick={30,35,40,45,50,55,60,65,70,75,80,85,90,95,100},
    legend style={
        at={(0.5,-0.4)},
    	anchor=south,
    	legend columns=3
	},
    ymajorgrids=true,
    xmajorgrids=true,
    grid style=dashed
]

\addplot[
    color=black,
    mark=triangle*
    ]
    coordinates{(3,62.18)(6,51.13)(12,49.5)(30,50)};
    
\addplot[
    color=black,
    mark=*
    ]
    coordinates{(3,56.2)(6,52.16)(12,53.43)(30,45.08)};

\addplot[
    color=black,
    mark=square*
    ]
    coordinates{(3,69.56)(6,63.01)(12,63.09)(30,53.06)};

\addplot[
    color=black,
    mark=triangle
    ]
    coordinates{(3,93.99)(6,89.72)(12,86.31)(30,74.83)};

\addplot[
    color=black,
    mark=o
    ]
    coordinates{(3,97.1)(6,95.66)(12,91.76)(30,78.81)};
    
\addplot[
    color=red,
    mark=square
    ]
    coordinates{(3,96.25)(6,93.18)(12,87.37)(30,77.33)};

\legend{\scriptsize{Delta}, 
\scriptsize{Theta},
\scriptsize{Alpha},
\scriptsize{Beta}, 
\scriptsize{Gamma}, 
\scriptsize{All Band}
}
    
\end{axis}

\end{tikzpicture}

    \caption{3-class classification accuracy on varying window size using AsMap+CNN features.}
    \label{fig:res1}
\end{figure}

\subsection{Two-Class Classification on DEAP}

On the DEAP dataset, valence and arousal classification accuracy were analyzed on different feature extraction methods. Table~\ref{table2} presents the valence classification accuracy obtained using different feature extraction methods on delta ($\delta$), theta ($\theta$), alpha ($\alpha$), beta ($\beta$), gamma ($\gamma$), and all frequency bands together (ALL BAND). In this experiment also, the window size was set to 3 s. The highest valence classification accuracy was achieved on ALL BAND using AsMap+CNN features, which was 95.45\%. However, the classification accuracy achieved by using DASM features on ALL BAND was very close to the accuracy using AsMap+CNN features. Further, the classification accuracy obtained by using DE, DASM, DCAU, and AsMap+CNN on ALL BAND was higher than that obtained with features on other frequency bands. In the $\beta$ and $\gamma$ bands, AsMap+CNN features generated the highest classification accuracy compared with other feature extraction methods. However, in the $\delta$, $\theta$, and $\alpha$ bands, the DE features yielded higher classification accuracy compared to other features. Table~\ref{table3} presents the arousal classification accuracy obtained using different feature extraction methods on delta ($\delta$), theta ($\theta$), alpha ($\alpha$), beta ($\beta$), gamma ($\gamma$), and all frequency bands together (ALL BAND). The highest arousal classification accuracy was achieved on ALL BAND using AsMap+CNN features, which was 95.21\%. However, the classification accuracy achieved using DCAU and DASM features on ALL BAND remained above 94\%. In comparison to valence classification, similar observations were made wherein the arousal classification accuracy obtained by using DE, DASM, DCAU, and AsMap+CNN on ALL BAND was higher than that obtained with features on other frequency bands. In the $\theta$, $\beta$, and $\gamma$ bands, AsMap+CNN features generated the highest classification accuracy compared with other feature extraction methods. However, in the $\delta$ and $\alpha$ bands, the DE features obtained higher classification accuracy compared to \mbox{other features. }
\end{paracol}
 \begin{table}[H] 
\caption{Valence classification accuracy obtained using different feature extraction techniques on  frequency bands.
\label{table2}}
\newcolumntype{C}{>{\centering\arraybackslash}X}
\begin{tabularx}{\textwidth}{CCCCCCC}
\toprule
\textbf{Method} & \boldmath{$\delta$} & \boldmath{$\theta$} & \boldmath{$\alpha$} & \boldmath{$\beta$} & \boldmath{$\gamma$} & \textbf{ALLBAND} \\
\midrule
DE & 80.44\% & 86.57\% & 86.46\% & 74.52\% & 80.20\% & 86.87\% \\
RASM & 56.71\% & 56.48\% & 57.60\% & 74.19\% & 70.69\% & 56.24\% \\
DCAU & 70.68\% & 74.84\% & 72.35\% & 74.07\% & 74.78\% & 93.20\% \\
DASM & 72.59\% & 78.61\% & 78.43\% & 78.48\% & 80.74\% & 95.08\% \\
AsMap+CNN & 79.61\% & 85.64\% & 86.15\% & 86.83\% & 86.57\% &95.45\% \\
\bottomrule
\end{tabularx}

\noindent{\footnotesize{ The window size was set to 3 s.}}
\end{table}
\unskip

 \begin{table}[H] 
\caption{Arousal classification accuracy obtained using different feature extraction techniques on  frequency bands.
\label{table3}}
\newcolumntype{C}{>{\centering\arraybackslash}X}
\begin{tabularx}{\textwidth}{CCCCCCC}
\toprule
\textbf{Method} & \boldmath{$\delta$} & \boldmath{$\theta$} & \boldmath{$\alpha$} & \boldmath{$\beta$} & \boldmath{$\gamma$} & \textbf{ALLBAND} \\
\midrule
DE & 82.01\% & 88.10\% & 87.78\% & 77.96\% & 80.65\% & 88.47\% \\
RASM & 57.55\% & 58.06\% & 64.08\% & 76.34\% & 74.49\% & 59.42\% \\
DCAU & 71.96\% & 75.90\% & 75.35\% & 75.27\% & 74.52\% & 94.60\% \\
DASM & 75.13\% & 81.03\% & 79.64\% & 79.31\% & 81.06\% & 94.17\% \\
AsMap+CNN & 81.38\% & 88.27\% & 87.24\% & 88.94\% & 89.00\% & 95.21\% \\
\bottomrule
\end{tabularx}
\noindent{\footnotesize{ The window size was set to 3 s.}}
\end{table}
\begin{paracol}{2}
\switchcolumn

The valence and arousal classification accuracy using AsMap+CNN on different frequency bands and window sizes are presented in Figures~\ref{fig:res2} and \ref{fig:res3}, respectively. Both the figures show a similar trend, where, with the increase in window size, the classification accuracy decreases. Using AsMap+CNN features on ALL BAND, the valence and arousal classification accuracy remained above 90\% for window sizes smaller than or equal to 12 s. 
The valence and arousal classification accuracy obtained showed linear degradation, and the accuracy remained above 68\% until a 30 s window size. Both Figures~\ref{fig:res2} and \ref{fig:res3} clearly show that AsMap+CNN features on ALL BAND together have greater discriminating ability compared to other bands for valence and arousal classification. 

\subsection{Four-Class Classification on DEAP}

In order to further test the capability of the AsMap+CNN feature extraction method, a four-class classification problem was formulated using the valence and arousal classes on the DEAP dataset. The four-class classification accuracy was also analyzed on other feature extraction methods. Table~\ref{table4} presents the four-class classification accuracy obtained using different feature extraction methods on delta ($\delta$), theta ($\theta$), alpha ($\alpha$), beta ($\beta$), gamma ($\gamma$), and all frequency bands together (ALL BAND). In this experiment also, the window size was set to 3 s. The highest classification accuracy of 93.41\% was achieved on ALL BAND using AsMap+CNN features. However, the classification accuracy achieved by using DASM features on ALL BAND was 92.23\%, which is close to the accuracy achieved using AsMap+CNN features. Similar to two-class classification, the four-class classification accuracy obtained using DE, DASM, DCAU, and AsMap+CNN on ALL BAND was higher than that obtained with features on other frequency bands. In the $\beta$ and $\gamma$ bands, AsMap+CNN features generated the highest classification accuracy compared with other feature extraction methods. However, in the $\delta$, $\theta$, and $\alpha$ bands, the DE features obtained higher classification accuracy compared to other features.
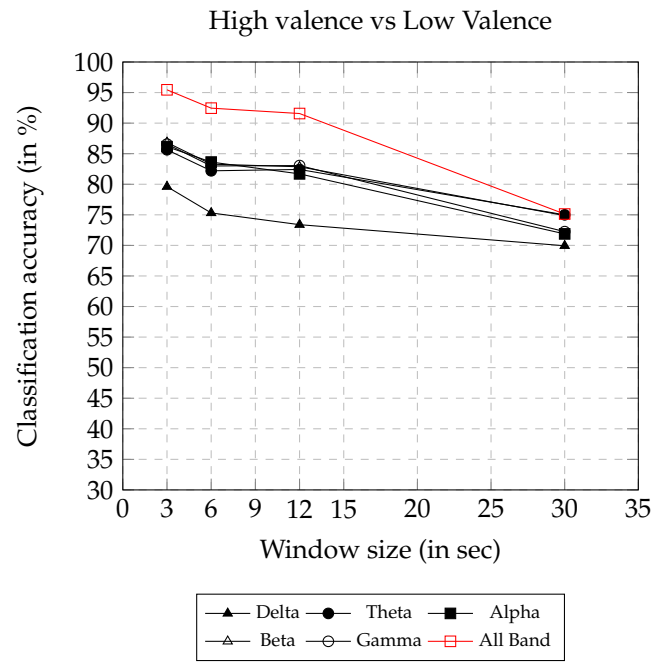
\begin{figure}[H]
\centering
    \begin{tikzpicture}
\begin{axis}[
    title={High valence vs Low Valence},
    xlabel={Window size (in sec)},
    ylabel={Classification accuracy (in \%)},
    xmin=0, xmax=35,
    ymin=30, ymax=100,
    xtick={0,3,6,9,12,15,20,25,30,35},
    ytick={30,35,40,45,50,55,60,65,70,75,80,85,90,95,100},
    legend style={
        at={(0.5,-0.4)},
    	anchor=south,
    	legend columns=3
	},
    ymajorgrids=true,
    xmajorgrids=true,
    grid style=dashed
]

\addplot[
    color=black,
    mark=triangle*
    ]
    coordinates{(3,79.61)(6,75.31)(12,73.39)(30,69.94)};
    
\addplot[
    color=black,
    mark=*
    ]
    coordinates{(3,85.64)(6,82.2)(12,82.42)(30,75)};

\addplot[
    color=black,
    mark=square*
    ]
    coordinates{(3,86.15)(6,83.63)(12,81.7)(30,71.88)};

\addplot[
    color=black,
    mark=triangle
    ]
    coordinates{(3,86.83)(6,83.29)(12,82.86)(30,74.88)};

\addplot[
    color=black,
    mark=o
    ]
    coordinates{(3,86.57)(6,82.99)(12,83.06)(30,72.25)};
    
\addplot[
    color=red,
    mark=square
    ]
   coordinates{(3,95.45)(6,92.45)(12,91.57)(30,75.13)};

\legend{\scriptsize{Delta}, 
\scriptsize{Theta},
\scriptsize{Alpha},
\scriptsize{Beta}, 
\scriptsize{Gamma}, 
\scriptsize{All Band}
}
    
\end{axis}

\end{tikzpicture}

    \caption{Valence classification accuracy on varying window size using AsMap+CNN features.}
    \label{fig:res2}
\end{figure}

\vspace{-12pt}

\begin{figure}[H]
\centering
    \begin{tikzpicture}
\begin{axis}[
    title={High Arousal vs Low Arousal},
    xlabel={Window size (in sec)},
    ylabel={Classification accuracy (in \%)},
    xmin=0, xmax=35,
    ymin=30, ymax=100,
    xtick={0,3,6,9,12,15,20,25,30,35},
    ytick={30,35,40,45,50,55,60,65,70,75,80,85,90,95,100},
    legend style={
        at={(0.5,-0.4)},
    	anchor=south,
    	legend columns=3
	},
    ymajorgrids=true,
    xmajorgrids=true,
    grid style=dashed
]

\addplot[
    color=black,
    mark=triangle*
    ]
    coordinates{(3,81.38)(6,76.66)(12,74.47)(30,69.13)};
    
\addplot[
    color=black,
    mark=*
    ]
    coordinates{(3,88.27)(6,85.09)(12,84.74)(30,78)};

\addplot[
    color=black,
    mark=square*
    ]
    coordinates{(3,87.24)(6,83.9)(12,81.36)(30,76.86)};

\addplot[
    color=black,
    mark=triangle
    ]
    coordinates{(3,88.94)(6,84.44)(12,83.63)(30,76.97)};

\addplot[
    color=black,
    mark=o
    ]
    coordinates{(3,89)(6,82.9)(12,84.17)(30,74.13)};
    
\addplot[
    color=red,
    mark=square
    ]
  coordinates{(3,95.21)(6,92.76)(12,89.61)(30,76.69)};

\legend{\scriptsize{Delta}, 
\scriptsize{Theta},
\scriptsize{Alpha},
\scriptsize{Beta}, 
\scriptsize{Gamma}, 
\scriptsize{All Band}
}
    
\end{axis}

\end{tikzpicture}

    \caption{Arousal classification accuracy on varying window size using AsMap+CNN features.}
    \label{fig:res3}
\end{figure}
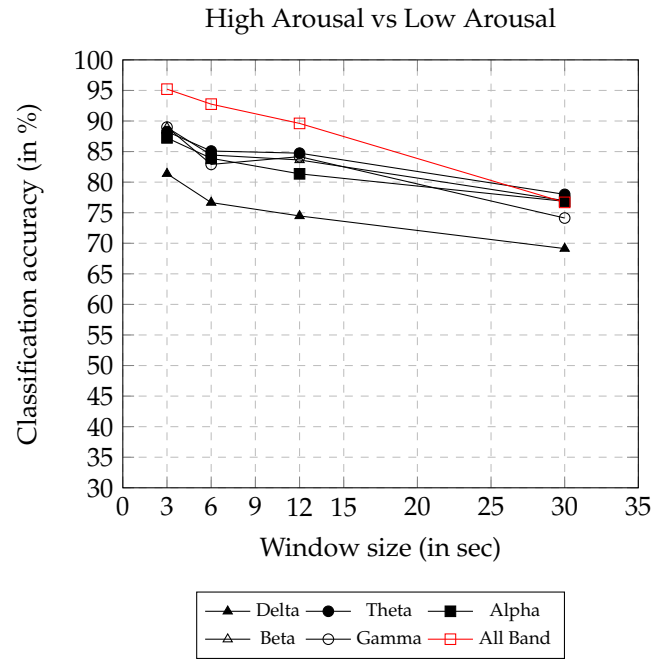
The four-class classification accuracy using AsMap+CNN on different frequency bands and window sizes are presented in Figure~\ref{fig:res4}. Similar to the observations in two-class and three-class classification, it was observed that the window size has a negative impact on classification accuracy. Using AsMap+CNN features on ALL BAND, the classification accuracy remained above 85\% for window sizes smaller than or equal to 12 s. 
However, the classification accuracy obtained on all frequency bands showed linear degradation, and the accuracy remained above 55\% until a 30 s window size. Figure~\ref{fig:res4} clearly shows that AsMap+CNN features on ALL BAND together have greater discriminating ability compared to other bands for complex classification problems having four classes. 
\end{paracol}
 \begin{table}[H] 
\caption{Four-class classification 
 accuracy obtained using different feature extraction techniques on  frequency bands.
\label{table4}}
\newcolumntype{C}{>{\centering\arraybackslash}X}
\begin{tabularx}{\textwidth}{CCCCCCC}
\toprule
\textbf{Method} & \boldmath{$\delta$} & \boldmath{$\theta$} & \boldmath{$\alpha$} & \boldmath{$\beta$} & \boldmath{$\gamma$} & \textbf{ALLBAND} \\
\midrule
DE & 70.23\% & 80.33\% & 80.89\% & 76.76\% & 79.31\% & 86.30\% \\
RASM & 30.97\% & 30.23\% & 47.15\% & 62.11\% & 59.11\% & 38.61\% \\ 
DCAU & 53.20\% & 62.71\% & 59.47\% & 58.87\% & 61.89\% & 90.48\% \\ 
DASM & 60.38\% & 69.65\% & 67.08\% & 67.57\% & 70.51\% & 92.23\% \\ 
AsMap+CNN & 67.86\% & 79.43\% & 79.15\% & 81.66\% & 82.16\% & 93.41\% \\ 
\bottomrule
\end{tabularx}
\noindent{\footnotesize{ The window size was set to 3 s.}}
\end{table}
\vspace{-16pt}

\begin{paracol}{2}
\switchcolumn
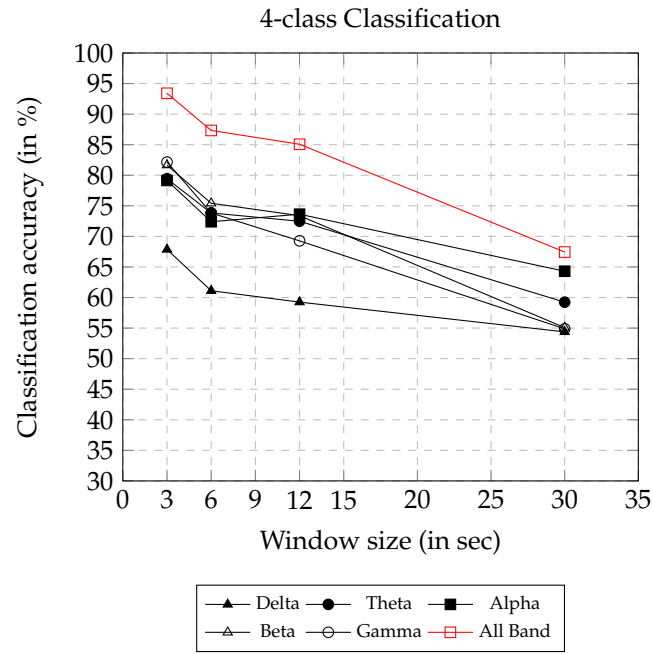
\begin{figure}[H]
\centering
    \begin{tikzpicture}
\begin{axis}[
    title={4-class Classification},
    xlabel={Window size (in sec)},
    ylabel={Classification accuracy (in \%)},
    xmin=0, xmax=35,
    ymin=30, ymax=100,
    xtick={0,3,6,9,12,15,20,25,30,35},
    ytick={30,35,40,45,50,55,60,65,70,75,80,85,90,95,100},
    legend style={
        at={(0.5,-0.4)},
    	anchor=south,
    	legend columns=3
	},
    ymajorgrids=true,
    xmajorgrids=true,
    grid style=dashed
]

\addplot[
    color=black,
    mark=triangle*
    ]
    coordinates{(3,67.86)(6,61.11)(12,59.25)(30,54.38)};
    
\addplot[
    color=black,
    mark=*
    ]
    coordinates{(3,79.43)(6,73.83)(12,72.48)(30,59.24)};

\addplot[
    color=black,
    mark=square*
    ]
    coordinates{(3,79.15)(6,72.4)(12,73.63)(30,64.3)};

\addplot[
    color=black,
    mark=triangle
    ]
    coordinates{(3,81.66)(6,75.42)(12,73.39)(30,55.08)};

\addplot[
    color=black,
    mark=o
    ]
   coordinates{(3,82.16)(6,73.83)(12,69.27)(30,54.88)};
    
\addplot[
    color=red,
    mark=square
    ]
   coordinates{(3,93.41)(6,87.33)(12,85.07)(30,67.44)};

\legend{\scriptsize{Delta}, 
\scriptsize{Theta},
\scriptsize{Alpha},
\scriptsize{Beta}, 
\scriptsize{Gamma}, 
\scriptsize{All Band}
}
    
\end{axis}

\end{tikzpicture}

    \caption{4-class classification accuracy on varying window size using AsMap+CNN features.}
    \label{fig:res4}
\end{figure}

\section{Discussion}
\label{sec:discussion}
In this experiment, the proposed hybrid feature extraction method (AsMap+CNN) outperformed other DE-based feature extraction methods in terms of classification accuracy. The proposed method was compared in competing scenarios where the window size was varied from 3 to 30 s. The accuracy of classification using the features was tested on different datasets and on a varying number of classes. On the DEAP dataset, AsMap+CNN features from all frequency bands achieved the highest valence and arousal classification accuracy of 95.45\% and 95.21\%, respectively. Further, experiments were conducted to increase the difficulty level by formulating a four-class classification problem on the DEAP dataset, and the highest classification accuracy of 93.41\% was achieved on ALL BAND using AsMap+CNN features. The highest classification accuracy of 97.10\% was achieved on the SEED dataset using AsMap+CNN features from the gamma band. One of the critical findings of this work is that AsMap+CNN on the gamma band generated more discriminative features than features from all bands together in classifying positive, negative, and neutral emotions on the SEED dataset. This indicates that emotional experience has a higher correlation with asymmetry in different brain regions on higher frequency bands. However, on the DEAP dataset, it was observed that features on all bands together provided higher classification accuracy than features on individual frequency bands. The DEAP dataset was prepared on 32 EEG channels, compared to 62 EEG channels  for the SEED dataset. The features generated have a lower spatial resolution, and features on individual bands do not provide classification accuracy above 90\%. Thus, with the power of CNN in learning hidden features, the classification accuracy increases by extracting hidden features from the AsMap on all bands. 

In contrast to other feature extraction methods, the AsMap captures the asymmetry among all the brain regions in a 2D vector. This work is the first attempt to generate AsMaps using DE features and feed them into a CNN for feature engineering, to the best of our knowledge. One of the limitations of this method is that the size of the AsMap increases with the increase in the number of EEG channels, which introduces a higher computational overhead on the CNN model. It was also observed that the classification accuracy shows linear degradation with the increase in window size. This is due to the fact that an increase in window size compromises the frequency resolution in STFT. Moreover, the window size is fixed while passing through the entire frequency spectrum. A viable solution to this is to use least-squares wavelet analysis (LSWA) or continuous wavelet transform (CWT) instead of STFT for more accurate estimation of frequencies and amplitudes~\cite{ghaderpour2021just,ghaderpour2021survey}. In LSWA or CWT, the window size decreases as the frequency increases, allowing one to capture the high-frequency components with short duration or with varying amplitude over time or frequency. The investigation of a frequency-dependent window length is subject to future work. The degradation in classification accuracy for large window sizes can also be attributed to the combination of more than one emotion feature in large windows. Investigation of the temporal features in the EEG data for a particular window can be a viable solution to the degradation in classification accuracy with an increase in \mbox{window size.}            

This work highlights the importance of hybrid feature extraction in emotion classification, as the accuracy of the classifier is directly dependent on the quality of features. The results demonstrate that the hybrid method of manual and automated feature extraction provides an advantage over the existing state-of-the-art feature extraction methods in emotion recognition systems using EEG. The proposed method's ability to classify discrete emotions in a valence--arousal coordinate space provides scope for advancement in EEG-based emotion recognition.

\section{Conclusions}
\label{sec:conclusion}

This work presented a deep learning approach for automated feature extraction for EEG-based emotion classification. 
As CNNs have shown potential in image classification, the DE features are transformed into a 2D feature vector called an AsMap. The automated features obtained using the AsMap on the CNN model provide the highest classification accuracy of 97.10\%, using a 3 s window size. The AsMap+CNN for feature extraction outperformed other feature extraction methods such as DE, DASM, RASM, and DCAU in terms of classification accuracy. The AsMap+CNN features capture the spatial correlation among different brain regions, thus resulting in higher classification accuracy. Results also indicated that the gamma band features give higher classification accuracy than other frequency bands on the SEED dataset. Further, experiments revealed that an increase in window size results in lower classification accuracy.



\vspace{6pt} 



\authorcontributions{Conceptualization: M.Z.I.A. and N.S.; methodology: M.Z.I.A.; Original draft preparation: M.Z.I.A.; Review and editing: E.G., N.S. and S.P.; supervision: N.S. All authors have read and agreed to the published version of the manuscript.}

\funding{This 
 research received no external funding.}

\institutionalreview{Ethical review and approval are not applicable for this study due to the use of public datasets.}
\informedconsent{Not applicable.}
\dataavailability{Not applicable}
\acknowledgments{The authors thank the editors and the reviewers for their time and constructive comments.}

\conflictsofinterest{The authors declare no conflicts of interest.}



\abbreviations{Abbreviations}{
The following abbreviations are used in this paper:\\

\noindent 
\begin{tabular}{@{}ll}
AsMap & asymmetric map\\
BCI & brain--computer interface\\
CNN & convolutional neural network\\
CWT & continuous wavelet transform \\
DASM & differential asymmetry\\
DCAU & differential caudality\\
DE & differential entropy\\
EEG & electroencephalogram \\
EMG & electromyogram\\
EOG & electrooculogram\\
GSR & galvanic skin response\\
HA & high arousal\\
HV & high valence\\
HVHA & high valence--high arousal\\
HVLA & high valence--low arousal\\
LSWA & least-squares wavelet analysis\\
LA & low arousal\\
LV & low valence\\
LVHA & low valence--high arousal\\
LVLA & low valence--low arousal\\
PSD & power spectral density\\
RASM & relative asymmetry\\
ReLU & rectified linear unit\\
RNN & recurrent neural network\\
SEED & SJTU Emotion EEG Dataset\\
STFT & short-time Fourier transform
\end{tabular}}


\end{paracol}
\reftitle{References}

\end{document}